\documentclass[journal]{IEEEtran}

\usepackage{cite}

\ifCLASSINFOpdf
\usepackage[pdftex]{graphicx}
\else
\usepackage[dvips]{graphicx}
\fi

\usepackage{amsmath}
\usepackage{amssymb}
\usepackage{multirow}

\newcommand{\etal}{\textit{et al}.}

\begin{document}

\title{Explainable High-order Visual Question Reasoning: A New Benchmark and Knowledge-routed Network}

\author{Qingxing Cao, Bailin Li, Xiaodan Liang and Liang Lin
	\IEEEcompsocitemizethanks{
		\IEEEcompsocthanksitem Q.Cao and X. Liang are with the School of Intelligent Systems Engineering, Sun Yat-sen University, China. B. Li is with DarkMatter AI Research. L. Lin is with the School of Data and Computer
		Science, Sun Yat-sen University, China
		\IEEEcompsocthanksitem Corresponding author:
		Xiaodan Liang (E-mail: xdliang328@gmail.com)}}
\markboth{IEEE Transactions on Image Processing, ~Vol.~X, No.~X, XXX}%
{Shell \MakeLowercase{\textit{et al.}}: Bare Demo of IEEEtran.cls for IEEE Journals}
\maketitle

\begin{abstract}
Explanation and high-order reasoning capabilities are crucial for real-world visual question answering with diverse levels of inference complexity (e.g., what is the dog that is near the girl playing with?) and important for users to understand and diagnose the trustworthiness of the system. Current VQA benchmarks on natural images with only an accuracy metric end up pushing the models to exploit the dataset biases and cannot provide any interpretable justification, which severally hinders  advances in high-level question answering. In this work, we propose a new HVQR benchmark for evaluating explainable and high-order visual question reasoning ability with three distinguishable merits: 1) the questions often contain one or two relationship triplets, which requires the model to have the ability of multistep reasoning to predict plausible answers; 2) we provide an explicit evaluation on a multistep reasoning process that is constructed with image scene graphs and commonsense knowledge bases; and 3) each relationship triplet in a large-scale knowledge base only appears once among all questions, which poses challenges for existing networks that often attempt to overfit the knowledge base that already appears in the training set and enforces the models to handle unseen questions and knowledge fact usage. We also propose a new knowledge-routed modular network (KM-net) that incorporates the multistep reasoning process over a large knowledge base into visual question reasoning. An extensive dataset analysis and comparisons with existing models on the HVQR benchmark show that our benchmark provides explainable evaluations, comprehensive reasoning requirements and realistic challenges of VQA systems, as well as our KM-net's superiority in terms of accuracy and explanation ability.
\end{abstract}


% \fi
\IEEEpeerreviewmaketitle

\begin{figure}[t]
	\centering
	\includegraphics[width=0.95\linewidth]{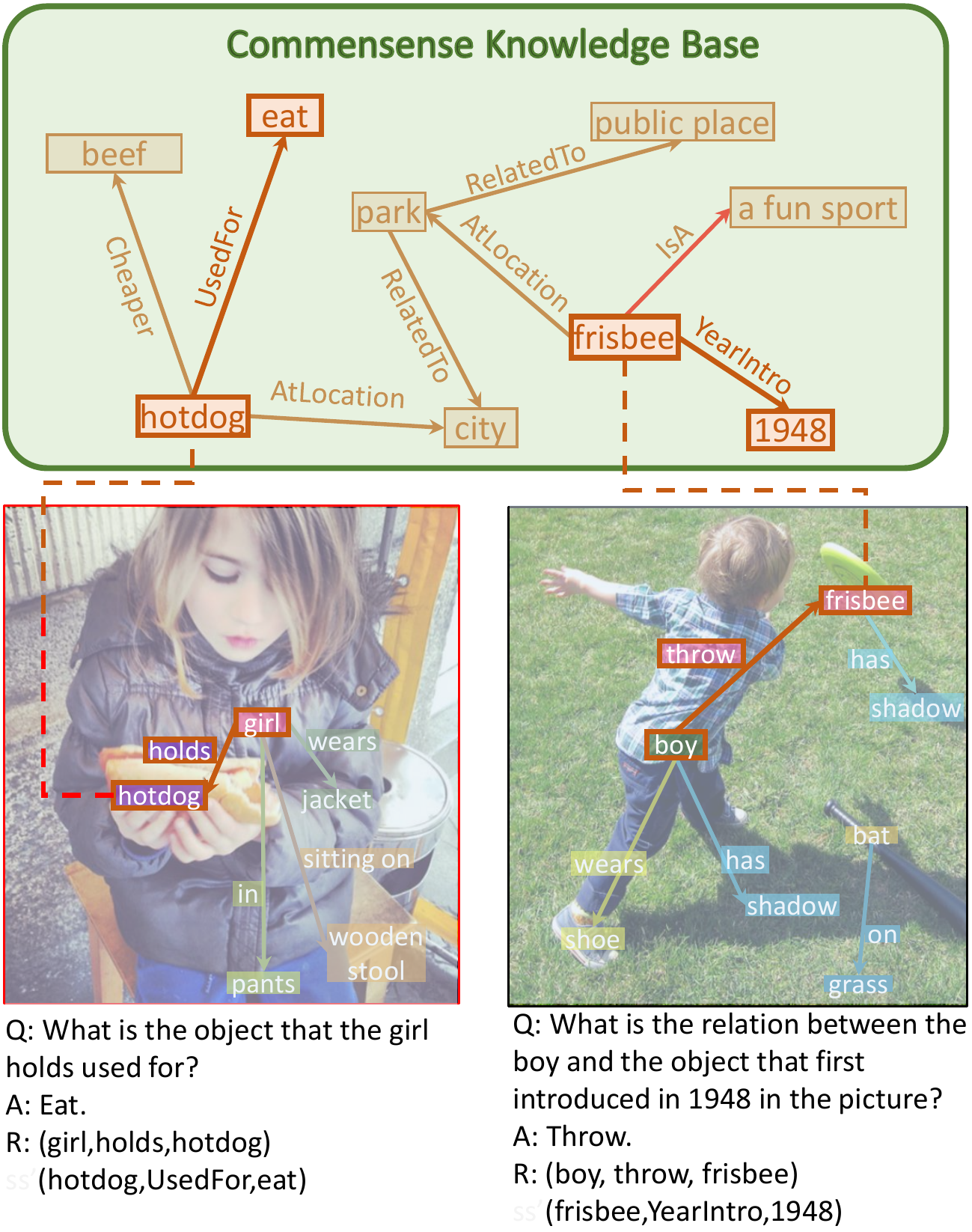}
	\caption{Examples of our HVQR dataset. Each question is generated by several relationship triplets selected from both scene graph annotations of the image and an external commonsense knowledge base. To answer the question, the VQA system is required to incorporate multistep reasoning over both of them. Furthermore, it provides an explainable evaluation metric to evaluate the reasoning process.}
	\label{fig:sample}
	
\end{figure}
\section{Introduction}    

Beyond the sole answer prediction metric that existing methods~\cite{mlb, film, san, buattention} and benchmarks~\cite{VQAv1, VQAv2, VG, V7W}  have focused on, the ultimate goal of visual question answering (VQA) is to train agents to be capable of understanding arbitrary questions with diverse levels of inference complexity (e.g., one-hop or multihop reasoning steps) and also provide an explainable diagnosis to improve the trustworthiness of the agent. A desirable agent that attempts to answer ``what is the object that the girl is holding used for'' should be able to understand the underlying entity relationships of the question, associate entities with visual evidence in the image, and reason correct answers by distilling mostly possible relations/attributes that may occur over the object of interest. We thus revisit current VQA research efforts that are enhanced by existing benchmarks to detect their critical limitations toward the long-standing cognitive VQA goal and then contribute a new explainable and high-order VQA benchmark to address these issues. 

Existing VQA datasets~\cite{VQAv1, VQAv2, VG, V7W} on natural-image scenarios often contain relatively simple questions and evaluate only the final answering prediction accuracy, lacking an evaluation of intermediate reasoning results and explainable justification about the system capability. Moreover, the human-annotated questions-answers suffer from high correlations, which enables the end-to-end networks~\cite{mlb, mfh, buattention, san, mfb, mcb} to achieve high performance by naively exploiting the dataset bias rather than co-reasoning on images and questions.  Some recent works address this problem by balancing the question-answer pairs. Johnson~\etal~\cite{CLEVR} proposed a synthetic dataset in which images and question-answer pairs are generated based on the given compositional layouts. However, a regular CNN with a fusion between image feature maps and question encoding~\cite{film} still has correctly answered these intricate and compositional questions. This makes it difficult to further evaluate the reasoning ability of the models on ~\cite{CLEVR}.

In this work, we propose a new high-order visual question reasoning (HVQR) benchmark to encourage a VQA agent to learn to perceive visual objects in the image, incorporate commonsense knowledge about the most common relationships among objects, and provide interpretable reasoning to the answer.%Editor: Please ensure that the intended meaning has been maintained in this edit.
Inspired by CLEVR~\cite{CLEVR} on synthetic datasets, our HVQR constructs the complex question-answer pairs based on the underlying layouts, consisting of one or two facts retrieved from a natural-image scene graph~\cite{VG} and external knowledge base~\cite{FVQA}. Moreover, we constrain the frequency of triplets from the knowledge base to prevent the black-box method from memorizing the correlation between question-answer pairs and knowledge triplets. Thus, the benchmark will enforce the model to perform multihop reasoning steps to handle unseen questions and evaluate the intermediate results, encouraging the generalization capability. Given an image, we first merge its scene graph and the common knowledge base to form an image-specific knowledge graph, and then we retrieve a route from the graph and ask  first-order or second-order questions based on the route. We illustrate the question-answer pairs of our HVQR in Figure~\ref{fig:sample}. Our constructed questions often contain multiple relation triplets and rare cases.

We also propose a knowledge-routed modular network (KM-net) to incorporate \textbf{explicit} multistep reasoning capability into the end-to-end network. First, our KM-net parses the question into the query layout. The query layout is the symbolic form of the retrieving and reasoning process on the knowledge graph. Then, given the query layout and knowledge base, KM-net attends the proper regions for entities and relations in the layout and performs multistep reasoning to achieve the final answer prediction.

We conduct extensive experiments under various levels of question complexity and comparisons with existing VQA benchmarks and state-of-the-art VQA models. The results show that the end-to-end methods often fail on questions that require external knowledge to reason over and that our KM-net outperforms all previous state-of-the-art VQA models on the HVQR benchmark. Meanwhile, experiments on the evaluation metric of explanation show the superior capability of explicit reasoning of our KM-net.

\begin{figure}[t]
	\centering
	\includegraphics[width=0.95\columnwidth]{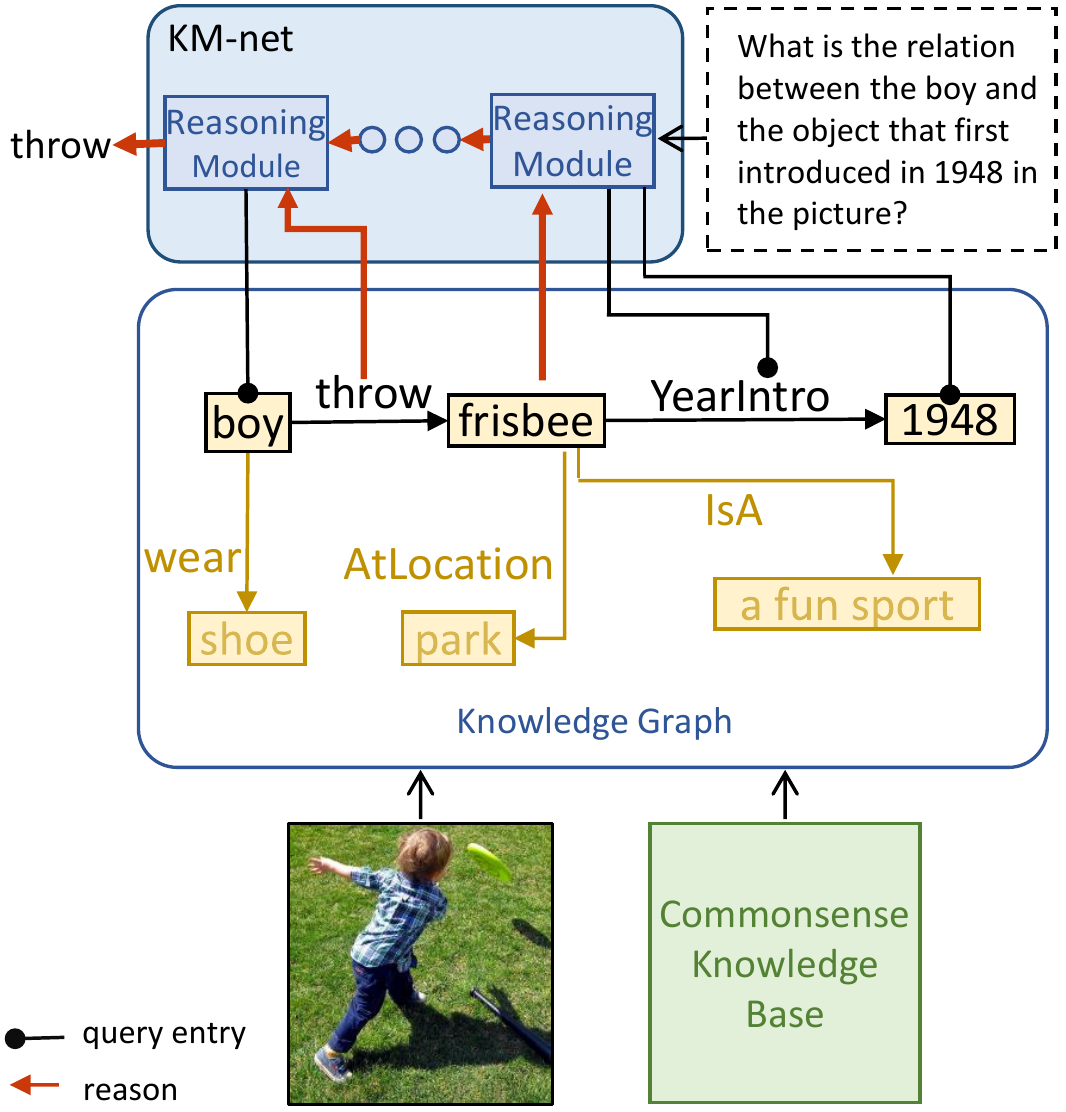}
	\caption{Given an image-question pair, our KM-net integrates a series of reasoning modules that enable incorporating the explicit reasoning over commonsense knowledge base into an end-to-end network. Given the query parsing layout, each reasoning module queries specific entities (black lines and boxes) and infers the possible reasoning entities/relationships (red lines and boxes) by reasoning in the knowledge base and visual attention over the image. Our KM-net thus performs multistep reasoning steps for traversing all relation triplets in the whole query layout and its corresponding visual evidence.}
	\label{fig:networkintro}
\end{figure}
\section{Related Works}
\textbf{Visual question answering.} Most of the methods on visual question answering tasks use an end-to-end model to predict the answer. The typical methods use a CNN to extract image features and an RNN to extract question features. Then, a bilinear fusion of these two features~\cite{mlb, mfh, mutan, film} is performed to predict the answer. Later, attention mechanisms\cite{san, coatt} following this pipeline achieved substantial improvements. Another line of work takes advantage of the structure inherently contained in text and images. NMN~\cite{nmn} and TbD-net~\cite{tbd} use handcrafted rules to generate network layouts, and ACMN~\cite{acmn} extends these methods to general questions using a dependency tree.

\textbf{Knowledge-based VQA.} In addition to the methods on general visual question answering, \cite{FVQA, Narasimhan_2018_ECCV} focus on the questions involving commonsense knowledge. \cite{FVQA} answers questions by learning query mappings to retrieve information in the knowledge base. \cite{Narasimhan_2018_ECCV} answers the questions by predicting the key triplet to the question. Both \cite{FVQA} and \cite{Narasimhan_2018_ECCV} are able to answer the questions of one-hop reasoning on the knowledge base.

\textbf{Datasets for VQA.} Datasets for visual question answering have been proposed in recent years. VQAv2~\cite{VQAv2} and Visual Genome~\cite{VG} are large-scale datasets on natural images. The questions in these datasets are not controlled by a knowledge graph, and most of them are not involved in complex reasoning. CLEVR~\cite{CLEVR} contains questions involving high-order visual reasoning. However, an end-to-end method~\cite{film} still answers the questions correctly due to the data bias of the synthetic images and questions. FVQA~\cite{FVQA} provides a VQA dataset along with a knowledge base. However, the questions only require first-order reasoning on the knowledge base. HVQR is a much larger dataset to involve high-order reasoning on both natural images and commonsense knowledge.

%%%%%%%%% BODY TEXT
\section{High-order Visual Question Reasoning Benchmark}

We first introduce a new HVQR benchmark that aims to push forward the research efforts on complex question requests with multiple inference steps (called high-order) and provides the first interpretable evaluation benchmark to enforce VQA models be explainable and possess a self-diagnosis capability. Compared to existing VQA datasets~\cite{VQAv1,  VQAv2, V7W, VG} on natural image scenarios, our HVQR can be distinguished from three aspects: 1) the questions often contain one or two relationship triplets, which requires the model to have the ability of multistep reasoning to predict plausible answers; 2)  we provide  an explicit  evaluation  on  a multistep  reasoning process that is constructed with image scene graphs and commonsense knowledge bases; and 3) each relationship triplet in the large-scale knowledge base only appears once among all questions, which poses challenges for existing networks that often attempt to overfit the knowledge base that already appears in the training set and enforces models to handle unseen questions and knowledge  facts. Detailed comparisons on the properties of our HVQR benchmark with existing datasets can be found in Table~\ref{tab:datasets}.

\begin{table}[t!]
	\centering
	\caption{Comparison of HVQR and previous datasets. Our HVQR dataset is the first large-scale (more than 100,000 QA pairs) dataset to require high-order reasoning on natural images.}
	\resizebox{\columnwidth}{!}{
		\begin{tabular}{cccccc}
			\hline
			Dataset & Natural image & High-order reasoning & Large scale & Knowledge base & Diagnostic \\
			\hline
			CLEVR~\cite{CLEVR} & & \checkmark & \checkmark & & \checkmark\\
			VQAv2~\cite{VQAv2} & \checkmark & & \checkmark & & \\
			FVQA~\cite{FVQA} & \checkmark & & & \checkmark & \checkmark \\
			Anab~\cite{anab} & \checkmark & \checkmark & & \checkmark & \checkmark \\
			\hline
			HVQR & \checkmark & \checkmark & \checkmark & \checkmark & \checkmark\\
			\hline
		\end{tabular}
	}
	
	\label{tab:datasets}
\end{table}

To generate a large-scale high-order VQR benchmark, we need to access the sophisticated understanding (all object relation triplets) of an image for generating diverse complex questions, that is, the image-wise scene graph~\cite{VG}, as well as a general commonsense knowledge base for injecting knowledge facts into the question reasoning. We thus construct our HVQR benchmark based on the scene graph annotations of the Visual Genome~\cite{VG} dataset, taking advantage of its comprehensive scene graph annotations in a large-scale image dataset on diverse and challenging scenarios. In this way, our HVQR benchmark finally contains 32,910 images from the Visual Genome~\cite{VG} dataset and a total of 157,201 question-answer pairs, of which 289,720 pairs are unique, which means they only appear once in the whole dataset. A total of 94,815 are used for training, 30,676 for validating and 31,710 for testing.

Specifically, given an image, we first extract facts from the image-specific knowledge graph by combining relation triplets from both the scene graph and the general knowledge base. Then, we compose these facts into a structural inference layout. Finally, we generate the question-answer pairs based on the inference layout and predefined templates. Questions are rewritten by a human to enhance the diversity and readability of the questions.

\subsection{Image-specific Knowledge Graph}
The image-specific knowledge graph that depicts the structured relation triplets appearing in one image is first built by integrating the image-wise scene graph and external commonsense knowledge base. 

First, the image-wise scene graph~\cite{VG} often describes the object relationships and attributes appearing in one image via a list of relation triplets, such as ``boy rides bike''. Given the image and its corresponding scene graph, we filter out the objects and relations that have the synset annotations and use the synset as their class label to reduce the semantic ambiguity. For instance, all objects that have the name ``bike'' or ``bicycle'' have the same label ``bicycle''.
To ensure that the generated questions require the VQA system to perform reasoning on commonsense knowledge, we enrich the knowledge set via a large-scale commonsense knowledge base from the common fact set. We extract a total of $193,449$ facts from WebChild~\cite{webchild}, ConceptNet~\cite{conceptnet}, and DBpedia~\cite{dbpedia} as the knowledge base set, following~\cite{FVQA}. Similar to~\cite{kvmn}, we denote a relationship or an entity as an entry.
Thus, each fact has three entries: a subject, a relationship, and an object.

We then merge the entries from the image-wise scene graph and general knowledge base into a large image-specific knowledge graph according to their synsets, which gathers many relation triplets to enable generating more realistic and complex question-answer pairs in HVQR.

\begin{table}[t!]
	\centering
	\caption{QA templates for generating rich high-order questions.}
	\resizebox{\columnwidth}{!}{
		\begin{tabular}{|c|c|c|c|c|}
			\hline
			Order & Qtype & Question Semantics & Answer & Reason\\
			\hline
			\multirow{3}{*}{1} & 0 & What is the relationship of $<$A$>$ and $<$B$>$? & $<$R$>$ & \multirow{3}{*}{(A, R, B)}\\
			& 1 & what is $<$A$>$ $<$R$>$? & $<$B$>$ &\\
			& 2 & what $<$R$>$ $<$B$>$? & $<$A$>$ &\\
			\hline
			\multirow{4}{*}{1} & 3 & What is the relation $<$of$>$ the object that $<$A$>$ $<$R1$>$ and $<$C$>$? & $<$R2$>$ & \multirow{4}{*}{}\\
			& 4 & What is the relation of $<$A$>$ and the object that $<$R2$>$ $<$C$>$? & $<$R1$>$ &(A, R1, B) \\
			& 5 & what $<$A$>$ $<$R1$>$ $<$R2$>$? & $<$C$>$ &(B, R2, C)\\
			& 6 & what $<$R1$>$ $<$R2$>$ $<$C$>$? & $<$A$>$ &\\
			\hline
		\end{tabular}
	}

	\label{tab:templates}
\end{table}  

\subsection{Question generation} 

\textbf{Question types.} The VQA task is proposed to promote and evaluate the model's comprehension of the visual scene. Typical questions in previous datasets~\cite{VQAv1, VQAv2, GaoYTM, V7W, VG} only contain single step inference, which cannot be further divided into any subquestions, such as ``What is on the desk?''. They are quite superficial, and most of them can be answered by black-box models. However, multihop reasoning questions facilitate models to think deeper
about what they see. Thus, our HVQR generates diverse multihop reasoning questions via structural compositions of a series of correlated elementary queries, such as ``What is the object that is on the desk used for?''. This enforces the agent to go through multistep structural reasoning (first identify the object on the desk, then infer its functionality) to answer the question.

We thus design 7 question composition types (see Table~\ref{tab:templates}), including different reasoning orders (single-step or multistep reasoning) and various knowledge involvement. Given several selected triplets from the extended image-wise knowledge graph, we first decide its \textit{Qtype}, then generate natural language QA pairs according to its corresponding question semantics. For instance, given a triplet ``(man, holds, umbrella)'' for \textit{Qtype 1},
``\textit{Que}: what is $<$A$>$ $<$R$>$ \textit{Ans}: $<$B$>$'', we generate a QA pair like ``\textit{Que}: what is the man holding \textit{Ans}: umbrella''. The reasoning order of a question can be decided according to the number of its given triplets. For example, ``(A, R, B)'' is designed for the first-order question, and ``(A, R1, B)-(B, R2, C)'' is designed for the second-order question. A question would be considered as ``KB-related'' if one of its triplets is sampled from the knowledge base; otherwise, it is ``KB-not-related''. For example, the question above is a ``KB-not-related'' question because ``(man, holds, umbrella)'' can be detected in the image. However, ``What is the usage of the object that the man is holding?'', which is converted from ``(man, holds, umbrella)-(umbrella, used for, keep out rain)'', is ``KB-related'' because ``(umbrella, used for, keep out rain)'' only appears in the knowledge base. For more examples, please refer to supplementary materials.

\textbf{Question collection.} We randomly sample several continuously linked triplets, such as ``(A, R1, B)-(B, R2, C)'' and convert them into natural language questions. Questions are first generated by 609 manually constructed templates. Considering the semantic rationality and uniqueness of the QA pairs, we use a functional program to walk through the image-specific knowledge graph to filter out invalid QA pairs. For example, a question that has more than one answer or has no answer would be considered an invalid question. To increase diversity and readability, questions are rewritten by human workers without changing their semantics, such as using synonyms or different sentence patterns. 

Noticing that the randomly sampled questions may lead to some dataset biases and poor model generalization, we also add more constraints to alleviate these challenges. To avoid the peak answer distribution, we restrict that each answer of a \textit{qtype} of all ``KB-related'' questions cannot appear more than 100 times. Moreover, to encourage diverse triplet distributions, we enforce that a specific knowledge fact in all questions cannot appear more than once for questions of \textit{qtype 2}, \textit{qtype 3} and \textit{qtype 5}.

\begin{figure}[t!]
	\centering
	\includegraphics[width=\columnwidth]{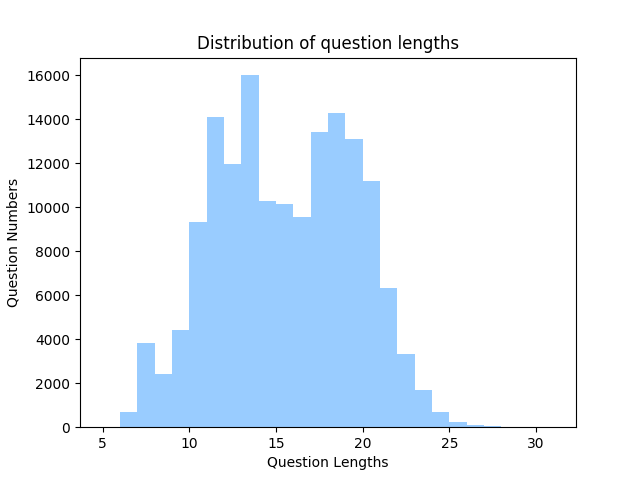}
	\caption{Distributions of the question lengths in HVQR.
	}
	\label{fig:qlengths}
\end{figure}
\begin{figure}[t!]
	\centering
	\includegraphics[width=\columnwidth]{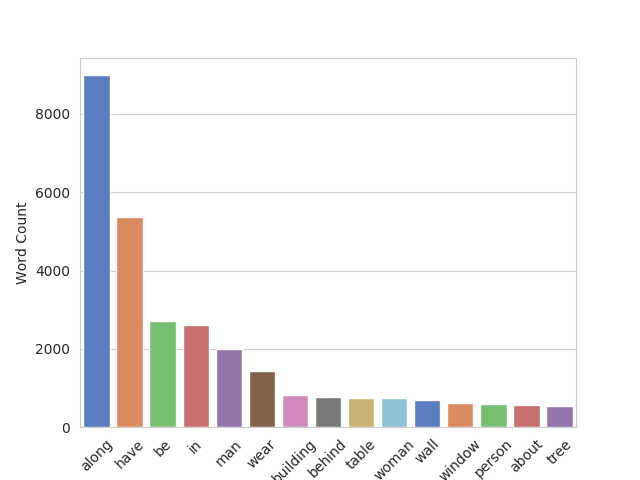}
	\caption{Top 15 words of the answers of the KB-not-related.
	}
	\label{fig:ans_nkb}
\end{figure}
\begin{figure}[t!]
	\centering
	\includegraphics[width=\columnwidth]{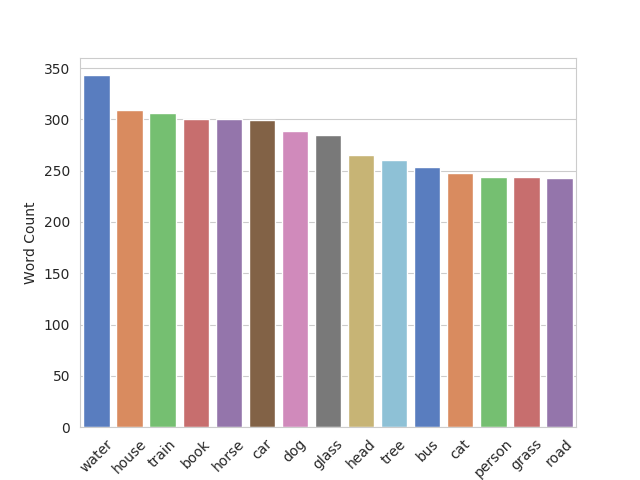}
	\caption{Top 15 words of the answers of the KB-related.
	}
	\label{fig:ans_kb}
\end{figure}

\subsection{Elementary queries}
As described in Section 4, there are 6 elementary queries. Here we detail the semantics of these queries.
Considering a triplet``(a, r, b)'', ``a'', ``r'' and ``b'' represent \textit{subject}, \textit{relationship} and \textit{object} respectively. Thus, the semantics of the queries can be denoted as below:
\begin{itemize}
	\item[-] \textbf{$Q_{ab\_I}$}: Given a \textit{subject} and an \textit{object}, return the \textit{relationship} between the \textit{subject} and the \textit{object} exploited in the image.
	\item[-] \textbf{$Q_{ar\_I}$}: Given a \textit{subject} and an \textit{relationship}, return the \textit{object} exploited in the image.
	\item[-] \textbf{$Q_{rb\_I}$}: Given an \textit{relationship} and an \textit{object}, return the \textit{subject} exploited in the image.
	\item[-] \textbf{$Q_{ab\_K}$}: Given a \textit{subject} and an \textit{object}, return the \textit{relationship} between the \textit{subject} and the \textit{object} in the knowledgebase.
	\item[-] \textbf{$Q_{ar\_K}$}: Given a \textit{subject} and an \textit{relationship}, return the \textit{object} in the knowledgebase.
	\item[-] \textbf{$Q_{rb\_K}$}: Given an \textit{relationship} and an \textit{object}, return the \textit{subject} in the knowledgebase. 
\end{itemize}

\begin{figure*}[t!]
	\centering
	\includegraphics[width=0.95\textwidth]{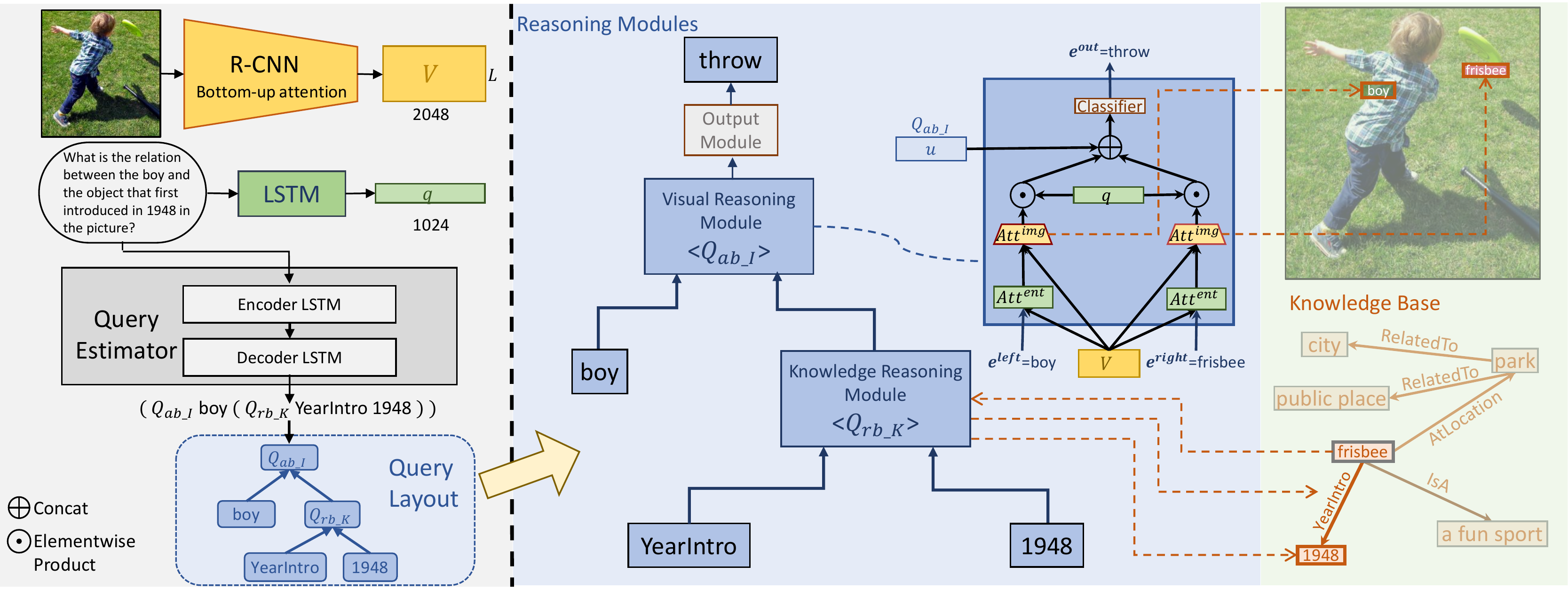}
	\caption{An overview of our KM-net. Given an image and question pair, we first parse the question into a structured query layout via a query estimator, where the question is represented by a triplet tree and embedded into a sentence embedding vector $\mathbf{q}$. The image is passed into a bottom-up attention R-CNN to obtain visual representation $\mathbf{V}$. Driven by the tree query layout, our KM-net routes the sequential reasoning modules to reason over the commonsense knowledge base and visual evidence in the image. For the query relation type $Q_{rb\_K}$, the knowledge reasoning module is first performed to reason candidate entities induced by this relation query, such as ``frisbee'' has the highest probability. Then, the visual reasoning module first attends visual features for all interested entities (e.g., boy or frisbee) and then fuses features of the question embedding $\mathbf{q}$, visual features of each entity and the query $Q_{ab\_I}$ to predict the open answers. 
	}
	\label{fig:network}
\end{figure*}

\subsection{Query Layout} HVQR also provides SQL-like query access to natural images, as mentioned in NMN \cite{nmn}, for further research. An elementary question can be mapped into an elementary query. A composition of subquestions can also be mapped into a composition of elementary queries. For example, ``What is the object that is on the desk used for?'' can be parsed into ``(${Q_{ar\_K}}$ ($Q_{rb\_I}$ on desk) UsedFor)'', where ``($Q_{rb\_I}$ R B)'' infers ``A'' in  triplet ``(A, R, B)'' and ``(${Q_{ar\_K}}$ A R)'' infers ``B''. We generate queries and answers as ground truths at each step of subqueries when converting triplets to natural language questions at the same time.

\subsection{Explainable evaluation metric} Existing VQA datasets \cite{FVQA, VQAv1, VQAv2, VG, CLEVR, V7W} only brute-force evaluate the VQA systems via the final answer accuracy, regardless of their interpretability and self-diagnosis capability. We argue that the desirable VQA systems that are truly applicable can not only predict the answer but also well explain the reasoning reasons to enable the proper justification and improve the trustfulness of the system. Guided by this principle, we introduce an explainable evaluation metric in our HVQR dataset. For each QA pair, the explainable evaluation metric calculates triplet precisions for each question, $recall=\frac{\#correct\_triplets}{\#predicted\_triplets}$ and the average recall of all QA pairs as the final recall.

\subsection{Dataset Statistics}

Our HVQR benchmark contains 32,910 images and 157,201 QA pairs, of which 138,560 are unique. They are split into the train, val and test splits at the ratio of 60\%, 20\% and 20\%, respectively (see Table~\ref{tab:data-all}). According to the reasoning steps, questions can be divided into {\bf first-order} and {\bf second-order} types of questions, which consist of 68,448 and 88,753, respectively.  According to the knowledge involvement of the questions, the questions can be divided into {\bf KB-related} and {\bf KB-not-related}, with 87,193 and 70,008, respectively. The lengths of the questions range from 4 to 24, 11.7 on average, and the longer questions correspond to questions that require more reasoning steps, as shown in Figure~\ref{fig:qlengths}.

\textbf{Answer distribution.} For the KB-not-related questions, the answer vocabulary size is 2,378 and shows a long-tailed distribution, as shown in Figure~\ref{fig:ans_nkb}. For the KB-related questions, the answer vocabulary size is 6,536, which is significantly greater than that of KB-not-related questions benefiting from many KB words. A total of 97\% of the answers in the val and test splits can be found in the train split. Because we have limited the maximal presence of each answer, the answers lie in a uniform distribution, as shown in Figure~\ref{fig:ans_kb}.

\textbf{Commonsense knowledge base.} Following~\cite{FVQA}, the knowledge base contains 193,449 knowledge triplets, 2,339 types of relations and 102,343 distinct entities. We restrict each triplet to only being present once for \textit{Qtype 2}, \textit{Qtype 3} and \textit{Qtype 5}, which enforces the machine to generate questions involving the diverse 45,550 knowledge triplets.

\begin{table}[t!]
	\centering
	\caption{Dataset statistics for different question types in HVQR.
	}
	\resizebox{\columnwidth}{!}{
		\begin{scriptsize}
			\begin{tabular}{|c|c|c|c|c|c|}
				\hline
				order & qtype & Train & Val & Test & Total \\
				\hline
				\multirow{3}{*}{1} & 0 & 8,133 & 2,730 & 2,698 & 13,561\\
				& 1 & 7,996 & 2,660 & 2,734 & 13,390\\
				& 2 & 25,171 & 7,982 & 8,344 & 41,497\\
				\hline
				\multirow{4}{*}{2} &3 & 9,477 & 2,931 & 3,131 & 15,539\\
				& 4 & 16,498 & 5,564 & 5,505 & 27,567\\
				& 5 & 10,482 & 3,291 & 3,518 & 17,291\\
				& 6 & 17,058 & 5,518 & 5,780 & 28,356\\
				\hline
			\end{tabular}
		\end{scriptsize}
	}
	
	\label{tab:data-all}
\end{table}

\section{Knowledge-routed Modular Network}
\subsection{Overview}
Given a question $X$ and an image $I$, our KM-net first parses the question into a query layout (consecutive queries) via a query estimator. Given the queries, two types of reasoning modules will retrieve factual knowledge or visual evidence from both the knowledge base and images, respectively. The final answer is predicted by the composition of elementary query results,
as shown in Figure~\ref{fig:network}.

\subsection{Query Estimator}
Our query estimator leverages the widely used sequence-to-sequence model in~\cite{iep}, which takes the word sequence of the question as input and predicts the sequence of query tokens. 

Each token in the predicted query sequence can be a separator (``('' or ``)''), name of entry (relationship or entity) or query symbol ($Q_{ab\_I}$, $Q_{ar\_I}$, $Q_{rb\_I}$, $Q_{ab\_K}$, $Q_{ar\_K}$, $Q_{rb\_K}$). Consider a triplet ``(A, R, B)'', $Q_{ab\_I}$ query for R, $Q_{ar\_I}$ query for B and $Q_{rb\_I}$ for A in the image. Additionally, $Q_{ab\_K}$, $Q_{ar\_K}$ and $Q_{rb\_K}$ query for the R, B and A, respectively, in the knowledge base. For example, ($Q_{ab\_I}$ boy frisbee) queries for the relation between boy and frisbee in the image. the answer should be throw.
Finally, the output tokens are parsed into a tree-structured layout using shift-reduce parser~\cite{shift}.

\begin{figure}[t!]
\label{fig:attent}
\centering
\includegraphics[width=1.0\columnwidth]{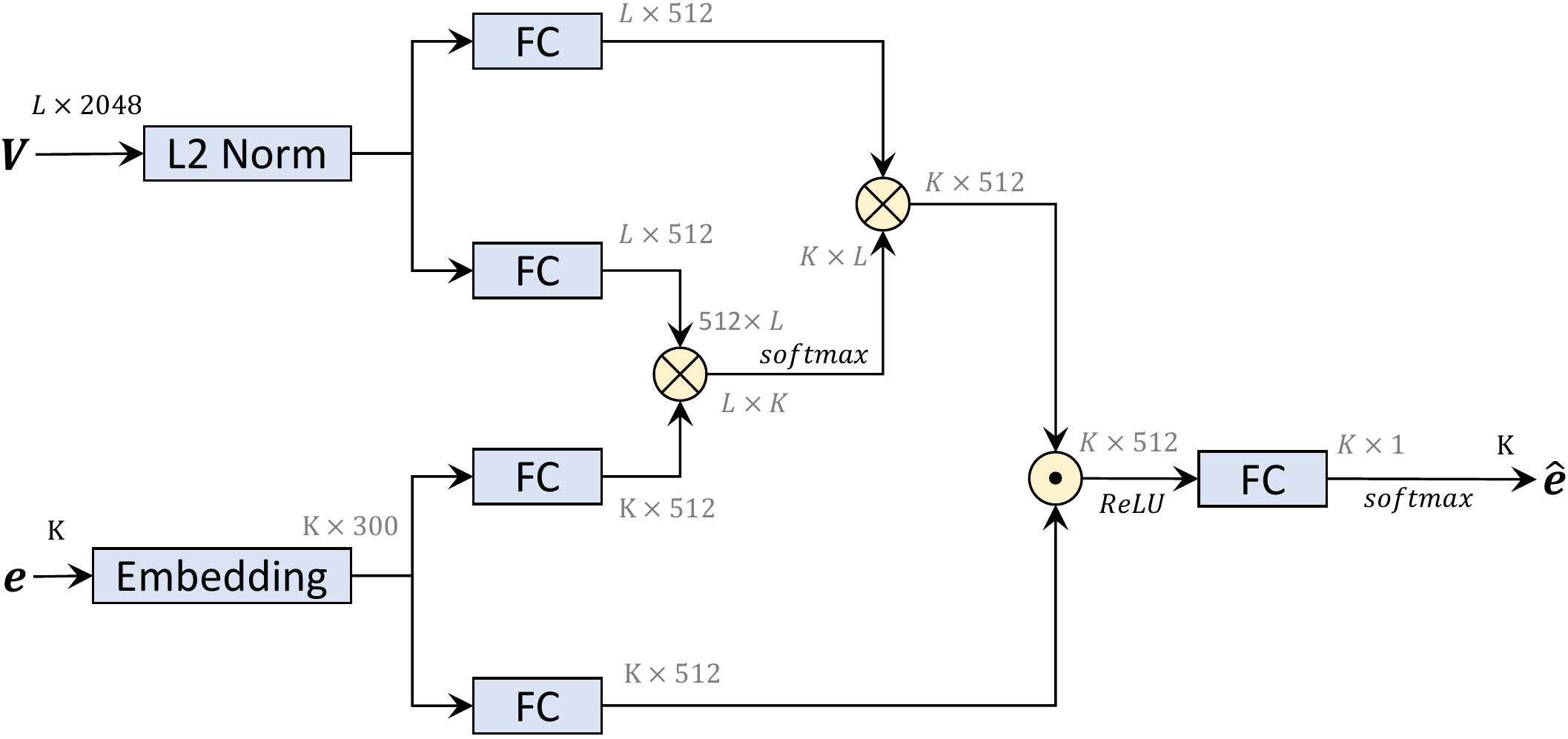}
\caption{a) The architecture details of entry attention module. We extract entry embeddings of the entries in the multi-hot vector $\mathbf{e}$, predict the attention weights over visual elements conditioned on the entry embeddings and perform attention over entries conditioned on the attended visual features to generate the weights vector of all entries $\mathbf{\hat{e}}$. 
}
\end{figure}

\subsection{Reasoning Modules}
The reasoning modules are routed by the query layout (bottom left part of Figure~\ref{fig:network}) from the query estimator. Bottom-up attention R-CNN~\cite{buattention} is a CNN that extracts features that provide region-specific information rather than grid-like feature maps. It is the state-of-the-art feature extractor in VQA tasks. We first extract the image features $\mathbf{V}$ using a bottom-up attention R-CNN and an LSTM similar to ~\cite{mfh} to extract question embedding vector $\mathbf{q}$. Then, we feed $\mathbf{q}$ and $\mathbf{V}$ into each visual reasoning module.

Each module takes the outputs of two child modules ($\mathbf{e^{left}}$ and $\mathbf{e^{right}}$) and performs the corresponding elementary query on the image $I$($Q_{ab\_I}$, $Q_{ar\_I}$, $Q_{rb\_I}$) or on the external knowledge base ($Q_{ab\_K}$, $Q_{ar\_K}$ or $Q_{rb\_K}$) to generate their own output $\mathbf{e^{out}}$ and passes it to its parent module.  $\mathbf{e}$ is a weight vector with the weights over all entries in the dataset. Finally, we feed the output of the root module into an output module, which is a two-layer MLP followed by a softmax layer, to predict the answer.

There are two types of modules in the modular network: \textit{knowledge reasoning module} and \textit{visual reasoning module}. At each node of the query layout, only one type is activated according to the \textit{query} type of the node. Every module with the same type shares their weights. During training, both the \textit{visual reasoning module} and output module are trained with cross-entropy loss. We now describe the details of the two types of modules.

\begin{figure}[t!]
	\label{fig:attimg}
	\centering
	\includegraphics[width=1.0\columnwidth]{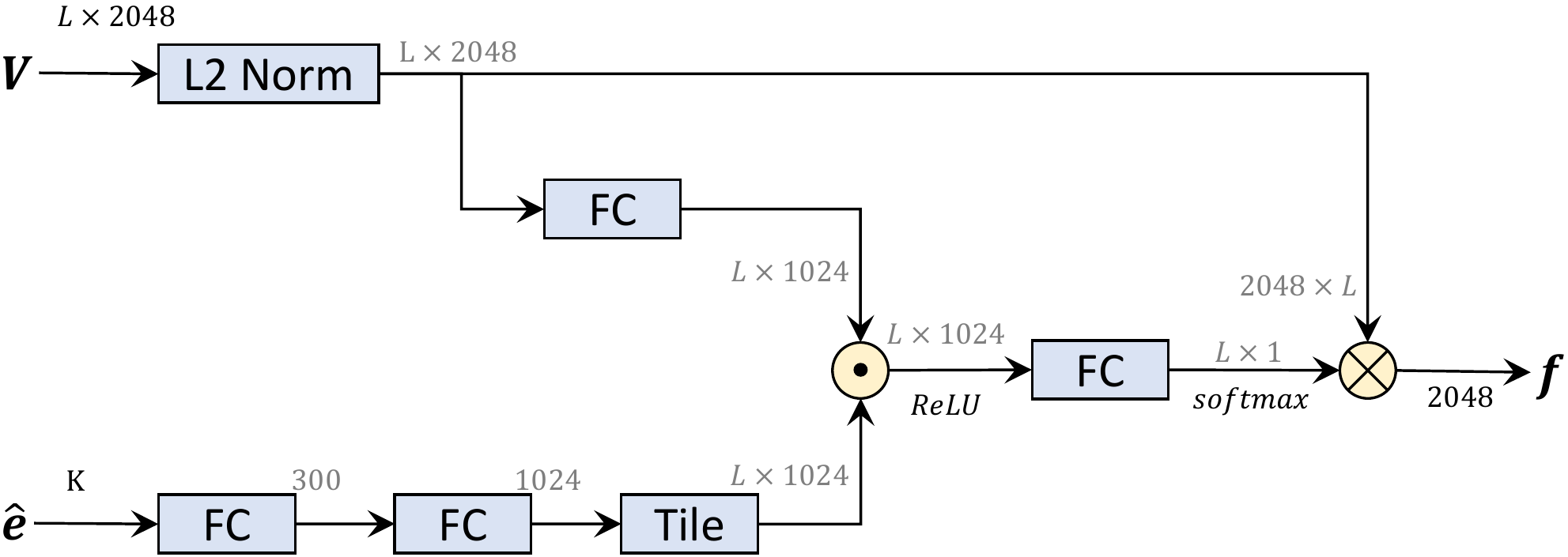}
	
	\caption{The architecture details of \textit{bilinear attention module}. We extract the attended entry embedding, fused the embedding with image feature and predict attention on the image to obtain attend visual feature $\mathbf{f}$.}
\end{figure}

\subsubsection{Knowledge Reasoning Module}
The knowledge reasoning module performs the elementary query with knowledge-related type explicitly on the external knowledge base.

When each knowledge reasoning module receives weight vectors ($\mathbf{e^{left}}$ and $\mathbf{e^{right}}$) from its two children, it first converts them into two entries by taking the entry (entity or relationship) with the maximal weight in $\mathbf{e^{left}}$ and $\mathbf{e^{right}}$. We denote the two taken entries as $s^{left}$ and $s^{right}$.

Then, the knowledge reasoning module performs operations on $s^{left}$ and $s^{right}$ according to the query type ($Q_{ab\_K}$, $Q_{rb\_K}$ or $Q_{ar\_K}$) of the module to generate a list of output entries $s^{out}$. Suppose that $s^{left}$ = UsedFor, $s^{right}$ = drink and the query is $Q_{rb\_K}$; the module would retrieve all entries that are used for drinking in the knowledge base. For example, ``water'' and ``milk'' are for drinking. Thus, $s^{out}$ = [water, milk]. Then, we convert $s^{out}$ into weight vector $\mathbf{e^{out}}$ simply using multi-hot encoding.

\begin{table*}[t!]
	\centering
	\caption{Comparison in terms of answer accuracy (\%) over the HVQR dataset. The performances include each \textit{qtypes}, KB-related or KB-not-related, each orders of reasoning level, baseline methods and KM-net.}
	\resizebox{\textwidth}{!}{
		\begin{tabular}{c|ccc|cccc|c|cccc|c}
			& \multicolumn{7}{c|}{KB-not-related} & \multicolumn{5}{c|}{KB-related} &\\
			& \multicolumn{3}{c|}{first-order} & \multicolumn{4}{c|}{second-order} & first-order & \multicolumn{4}{c|}{second-order} & \\
			Method & 0 & 1 & 2 & 3 & 4 & 5 & 6 & 2 & 3 & 4 & 5 & 6 & Overall\\
			\hline
			\hline
			Q-type mode & 36.19 & 2.78 & 8.21 & 33.18 & 35.97 & 3.66 & 8.06 & 0.09 & 0.00 & 0.18 & 0.06 & 0.33 & 8.12\\
			LSTM & 45.98 & 2.79 & 2.75 & 43.26 & 40.67 & 2.62 & 1.72 & 0.43 & 0.00 & 0.52 & 1.65 & 0.74 & 8.81 \\
			FiLM~\cite{film} & 52.42 & 21.35 & 18.50 & 45.23 & 42.36 & 21.32 & 15.44 & 6.27 & 5.48 & 4.37 & 4.41 & 7.19 & 16.89\\
			MFH~\cite{mfh} & 43.74 & 28.28 & 27.49 & 38.71 & 36.48 & 20.77 & 21.01 & 12.97 & 5.10 & 6.05 & 5.02 & 14.38 & 19.55\\
			GRU+Bottom-up attention~\cite{buattention} & \textbf{56.42} & \textbf{29.89} & \textbf{28.63} & \textbf{49.69} & \textbf{43.87} & \textbf{24.71} & \textbf{21.28} & 11.07 & 8.16 & 7.09 & 5.37 & 13.97 & 21.85\\
			\hline
			KM-net & 53.42 & 28.68 & 26.43 & 41.52 & 43.32 & 24.59 & 20.04 & \textbf{26.98} & \textbf{11.17} & \textbf{10.82} & \textbf{15.36} & \textbf{15.74} & \textbf{25.19}\\
			
		\end{tabular}
	}
	
	\label{tab:acc}
\end{table*}

\subsubsection{Visual Reasoning Module}
Similar to the knowledge reasoning module, the visual reasoning module also takes two weight vectors as input and outputs one weight vector. However, it performs learnable operations over an image rather than retrieving entries in the knowledge base.

Because the operations on $\mathbf{e^{left}}$ and $\mathbf{e^{right}}$ are the same, we only describe operations on $\mathbf{e^{left}}$.
Given the input entity vector $\mathbf{e}$, we feed it into \textit{entry attention module} $\mathbf{Att^{ent}}$ used in Question-Image Co-Attention~\cite{coatt} to generate attention weights (denoted $\mathbf{\hat{e}}$) over all entries, as shown in Figure~\ref{fig:attent}. We first multiply $\mathbf{e}$ with all the entry embeddings in the lookup table to extract entry embeddings of the entries activated in the multi-hot vector. Then we perform a bilinear fusion to predict the attention weights of each entries over the visual elements. The attended features corresponding to each entry is generated by calculating the weighted sum of the visual elements. Finally, we perform a bilinear fusion between entry embeddings and the attended features to predict the weights vector of all entries, denoted $\mathbf{\hat{e}}$. This can be view as a refinement of the input weight vector (see $\mathbf{Att^{ent}}$ in Figure~\ref{fig:network}). If $\mathbf{e}$ is generated by a preceding visual reasoning module, we directly set $\mathbf{\hat{e}}$ = $\mathbf{e}$ as it is already a weight vector over all entries.

We then use $\mathbf{\hat{e}^{left}}$ to generate attention feature $\mathbf{f^{left}}$ over the image feature $\mathbf{V}$ using a bilinear attention module similar to~\cite{blatt}. We first pass the weight vector $\mathbf{\hat{e}}$ through a fully-connected layer, which is equivalent to the weighted sum of all the entry embeddings. Then we perform bilinear fusion between the the weighted sum and image feature to predict attention weight over the image. Finally, we calculate the weighted sum of the image elements to generated the feature $\mathbf{f}$, as shown in Figure~\ref{fig:attimg}.

At last, we fuse the embeddings of entries, question and image attention features using element-wise product to generate the fused feature $\mathbf{m^{left}}$. In general, the operations described in this paragraph can be formulated as:
\begin{equation}
\begin{aligned}
\mathbf{\hat{e}^{left}} &= \mathbf{Att^{ent}}(\mathbf{e^{left}}, \mathbf{V}), \\
\mathbf{f^{left}} &= \mathbf{Att^{img}}(\mathbf{\hat{e}^{left}}, \mathbf{V}) \\
\mathbf{m^{left}} &= \mathbf{ReLU}((\mathbf{W^H}\mathbf{E}\mathbf{\hat{e}^{left}}) \odot (\mathbf{W^G}\mathbf{q}) \odot \mathbf{f^{left}}) \\
\end{aligned}
\label{eq:vrm}
\end{equation}
where $\odot$ is the elementwise product, $\mathbf{W^H}$ and $\mathbf{W^G}$ are learnable transformation matrices, and $\mathbf{E}$ is the lookup table of all entries in the dataset.

$\mathbf{m^{right}}$ can be generated by $\mathbf{e^{right}}$ with the same operations that are performed on $\mathbf{e^{left}}$. We concatenate $\mathbf{m^{left}}$, $\mathbf{m^{right}}$ and query embedding vector $\mathbf{u}$ and feed it into a two-layer fully connected layer to generate output vector $\mathbf{e^{out}}$, which is the output of this module.

\begin{figure}[t!]
	\includegraphics[width=\columnwidth]{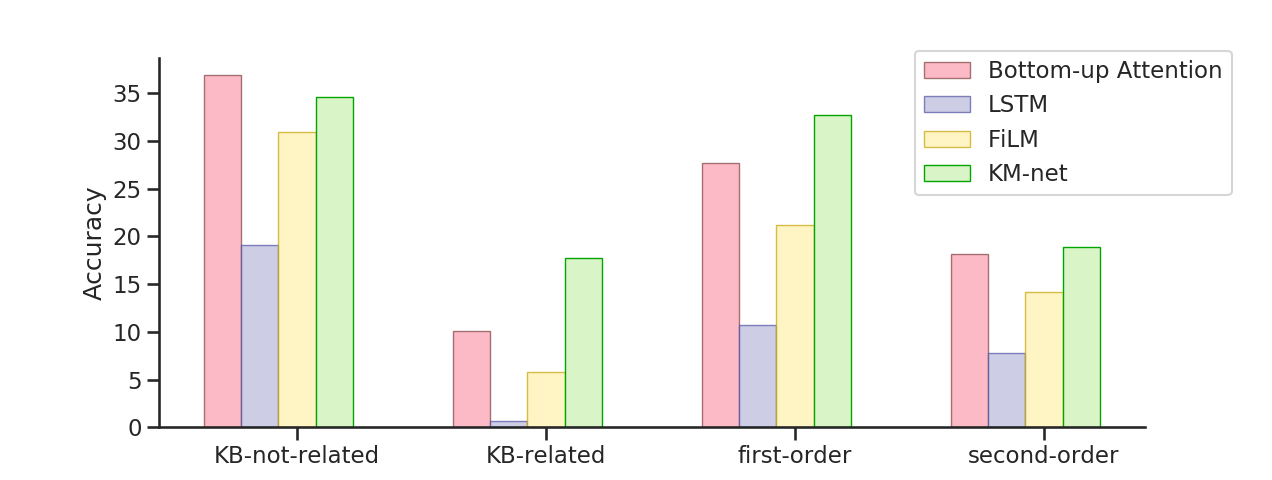}
	\caption{Comparison of different representative methods.
	}
	\label{fig:compare}
\end{figure}

\begin{figure*}[t!]
	\includegraphics[width=\textwidth]{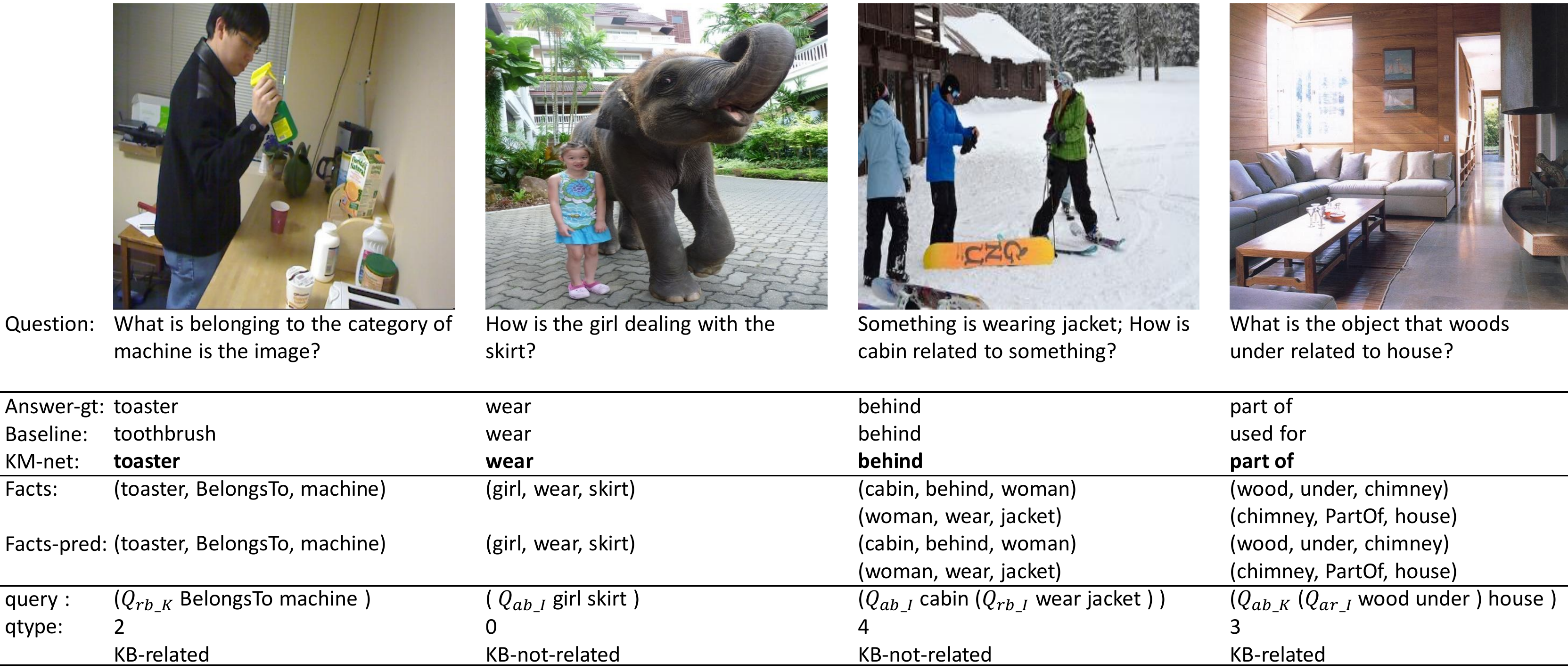}
	\caption{Examples of the results of baseline (bottom-up attention) and KM-net (Answer-gt: ground-truth answer, Fact: ground-truth supporting facts of the answer, Facts-pred: supporting facts predicted by KM-net).
	}
	\label{fig:dataviz}
\end{figure*}

\section{Experiments}
We first perform experiments to compare our KM-net with state-of-the-art methods on our proposed HVQR benchmark. Then, we evaluate the reasoning ability or our proposed model by comparing the retrieved knowledge with the ground-truth supporting facts. The implementation details of our model are given in the supplemental materials.

FVQA~\cite{FVQA, Narasimhan_2018_ECCV} and Anab~\cite{anab} are not evaluated on our benchmark since these methods heavily rely on the dataset-specific query annotations format. It will need considerable modification on their model to train their method on our datasets, which results in great difficulties in maintaining the accuracy of their model. 

\subsection{Answering Accuracy}
\textbf{Q-type mode.} This baseline method simply takes the most frequent answer in the training set as its output for each question type.

\textbf{LSTM.} We used an LSTM to encode the whole questions, and then the question embedding is fed into a two-layer MLP to predict the final answer. 
The above two baseline results are to show the degree of bias in our proposed dataset, similar to other works on VQA datasets~\cite{VQAv1, CLEVR, FVQA}

\textbf{GRU+Bottom-up attention.} We evaluate the winning method~\cite{tips} of the 2017 VQA challenge. This method used a GRU to encode the questions and extracted the image feature using a Fast-RCNN trained on the Visual Genome dataset. For simplicity, we extract the features from the top-$36$ object proposals, as provided by ~\cite{buattention}. Additionally, we perform multimodel bilinear pooling~\cite{mlb} and attention mechanism on the image feature and question embedding. The attended feature is passed through MLPs to predict the answer.

\textbf{MFH.} This method~\cite{mfh} is the runner-up of the 2018 VQA challenge. Similar to ~\cite{tips}, an image is encoded by a CNN with a bottom-up CNN, and the question is encoded with an LSTM. It then performs attention and question and image separately. The attention features are fused using stacks of bilinear fusion blocks~\cite{mfh}, which is improved based on~\cite{mlb}. The answer is predicted by passing the fused feature through an MLP.

\textbf{FiLM}~\cite{film} This method fuses the question embedding into different layers of a convolutional network with elementwise multiplication and addition. It has achieved high accuracy on the CLEVR~\cite{CLEVR} dataset, which is composed of questions that need compositional reasoning ability to correctly answer.

\textbf{KM-net.} We evaluate our proposed method, KM-net, as described above. Because our model consists of a query estimator and a modular network, we evaluate it with three settings: 1) FULL, given ground-truth query layout and scene graph annotations for both training and testing, the answer is generated by executing programs. 2) QUERY, the same as 1) but the ground-truth query layout is not given when testing. 3) KM-net w/ layout, both ground-truth query layout and scene graph annotation are provided for training; when testing, the module guided by the ground-truth layout should predict the answer by the trained modules.%Editor: Please ensure that the intended meaning has been maintained in this edit.
4) KM-net, the same as 3) when training, but the model should predict the answer guided by predicted query layout and the trained module when testing. Table~\ref{tab:acc} shows the answer accuracies of different works on HVQR.

As shown, our method outperforms all the previous methods in terms of overall accuracy. Concretely, the knowledge reasoning module in our KM-net can retrieve knowledge in the knowledge base, and it outperforms previous methods by a large margin on KB-related questions (particularly \textit{qtype 2} and \text{qtype 5}). LSTM and Q-type mode perform extremely inferior because they only access questions and predict the answer blindly. FiLM almost doubles the accuracy of LSTM because it takes the image as input, which provides more information to predict the answer. MFH and GRU+Bottom-up attention perform slightly better than FiLM because they use a bottom-up attention R-CNN to extract the feature. Figure~\ref{fig:dataviz} shows some examples of the predicted results of the representative baseline (GRU+Bottom-up attention) and KM-net.

\textbf{Reasoning and Commonsense Knowledge.} As shown in Figure~\ref{fig:compare}, all the methods perform worse on second-order questions than on first-order questions. This result shows that high-order reasoning is challenging. Our method outperforms all the methods by a large margin, which almost doubles the accuracy of the second-best method. It shows that our method can  incorporate commonsense knowledge while reasoning.

\begin{figure}[t]
	\includegraphics[width=\columnwidth]{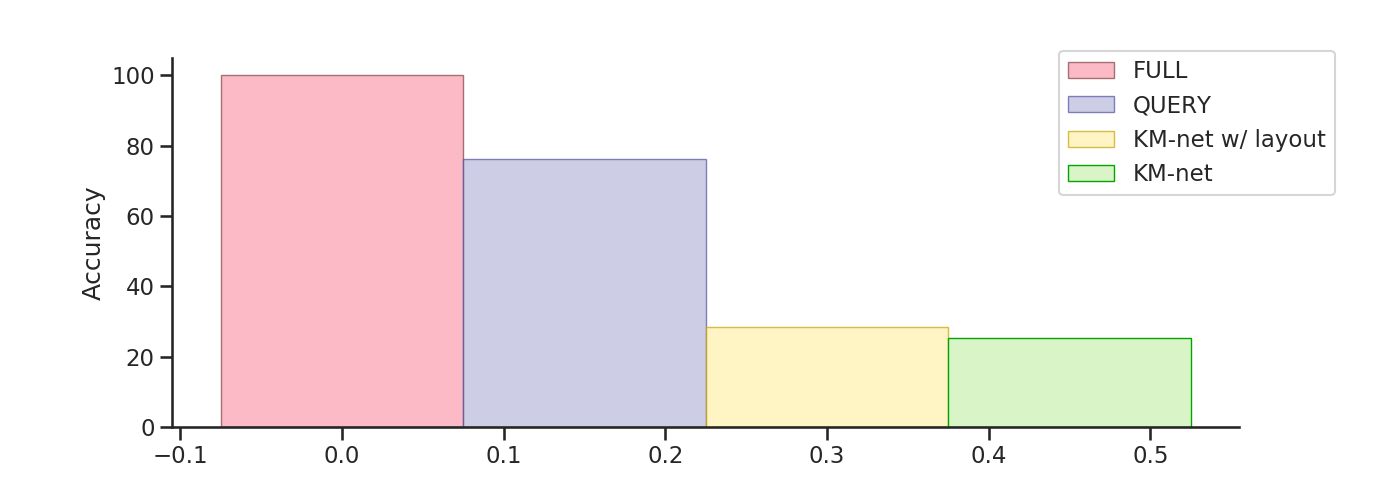}
	\caption{Ablation studies. Accuracy (\%) of the four settings: FULL, QUERY, KN-net w/ layout and KM-net.
	}
	\label{fig:ablation}
\end{figure}

\textbf{Ablation studies.} We conduct experiments of the four settings described above to validate the reliability of the HVQR dataset (see Figure~\ref{fig:ablation}). FULL achieves 100\%, which shows the correctness of the HVQR dataset: all the questions have exactly one answer, and all of them can be answered with the query. QUERY achieves 75.87\%, which shows the accuracy of the query prediction. The performances of KM-net w/ layout and KM-net drop seriously (28.05\% and 25.19\%) when the scene graph annotation is not given while testing. This result indicates that scene understanding and reasoning is challenging. KM-net performs slightly worse than KM-net w/ layout as the query layout is not given while testing.

\begin{table}[t!]
	\centering
	\caption{Results of evaluation metric of explanation, which is the average recall of the predicted supporting triplets.}
	\resizebox{\columnwidth}{!}{
		\begin{tabular}{c|c|c|c}
			Method & KB-not-related & KB-related & Overall \\
			\hline
			\hline
			KM-net w/ layout & 28.78 & 27.38 & 28.00\\
			KM-net & 25.71 & 20.97 & 23.08\\
		\end{tabular}
	}
	
	\label{tab:reason}
\end{table}

\subsection{Evaluation metric of explanation} In addition to answer accuracy, we also evaluate the reasoning process by evaluating the predicted supporting triplets. Concretely, we take the inputs ($\mathbf{e^{left}}$ and $\mathbf{e^{right}}$) and outputs ($\mathbf{e^{out}}$) of each module and formulate predicted triplets according to query type. If $e$ is more than one word, each word in $e$ would be taken to formulate a triplet separately. For example, $\mathbf{e^{left}}$=[apple, orange], $\mathbf{e^{right}}$=[on], $\mathbf{e^{out}}$=[plate], and \textit{query}=$Q_{ar\_I}$. Then, the predicted supporting triplets are (apple, on, plate) and (orange, on, plate). With the predicted triplets, we are able to evaluate the reasoning process with the evaluation metric described above (see Table~\ref{tab:reason}). As shown, the precisions of our model distributed similarly to the accuracy, which shows that our model can provide supporting facts while predicting answers. The metric provides a diagnostic approach to evaluate the methods.

\section{Conclusion}
In this paper, we propose a new visual question reasoning benchmark, which can evaluate a method's ability on high-order visual question reasoning. We also propose a novel knowledge-routed modular network that performs multistep reasoning explicitly by incorporating visual knowledge and commonsense knowledge. The experiments shows the superiority of our KM-net in terms of both accuracy and explainability.

\ifCLASSOPTIONcaptionsoff
\newpage
\fi
\bibliographystyle{IEEEtran}
\bibliography{egbib}
\newpage
\begin{IEEEbiography}[{\includegraphics[width=1in,height=1.25in,clip,keepaspectratio]{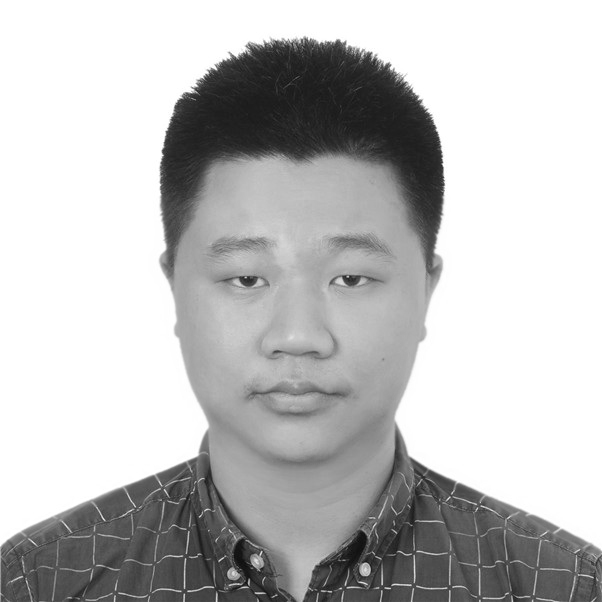}}]{Qingxing Cao}
	Qingxing Cao is currently a postdoctoral researcher in the School of Intelligent Systems Engineering at Sun Yat-sen University, working with Prof. Xiaodan Liang. He received his Ph.D. degree from Sun Yat-Sen University in 2019, advised by Prof. Liang Lin. His current research interests include computer vision and visual question answering.
\end{IEEEbiography}
\begin{IEEEbiography}[{\includegraphics[width=1in,height=1.25in,clip,keepaspectratio]{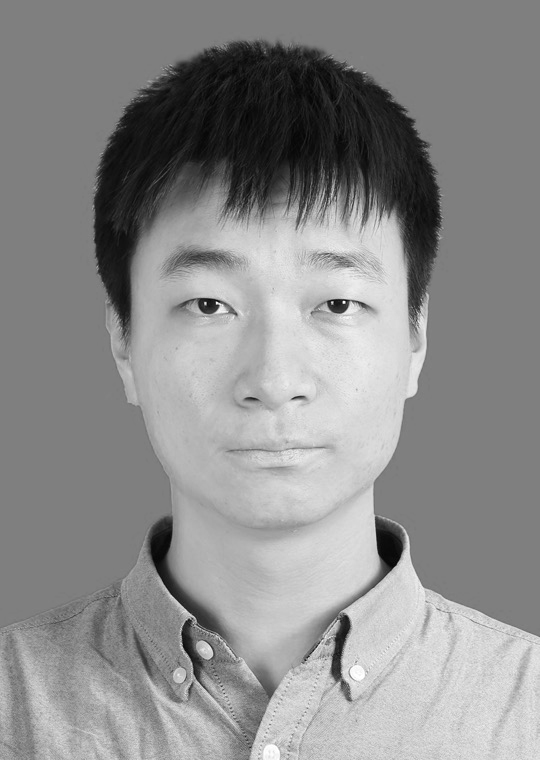}}]{Bailin Li}
	Bailin Li received his B.E. degree from Jilin University, Changchun, China, in 2016, and the M.S. degree at Sun Yat-Sen University, Guangzhou, China, advised by Professor Liang Lin. He currently leads the model optimization team at DMAI. His current research interests include visual reasoning and deep learning (e.g., network pruning, neural architecture search).
\end{IEEEbiography}
\begin{IEEEbiography}[{\includegraphics[width=1in,height=1.25in,clip,keepaspectratio]{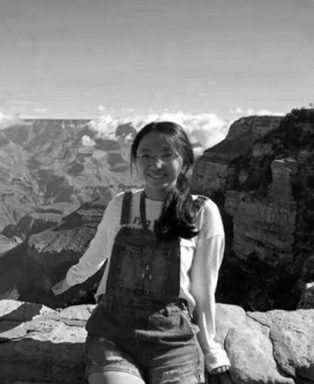}}]{Xiaodan Liang}
	Xiaodan Liang is currently an Associate Professor at Sun Yat-sen University. She was a postdoc researcher in the machine learning department at Carnegie Mellon University, working with Prof. Eric Xing, from 2016 to 2018. She received her PhD degree from Sun Yat-sen University in 2016, advised by Liang Lin. She has published several cutting-edge projects on human-related analysis, including human parsing, pedestrian detection and instance segmentation, 2D/3D human pose estimation and activity recognition.
\end{IEEEbiography}
\begin{IEEEbiography}[{\includegraphics[width=1in,height=1.25in,clip,keepaspectratio]{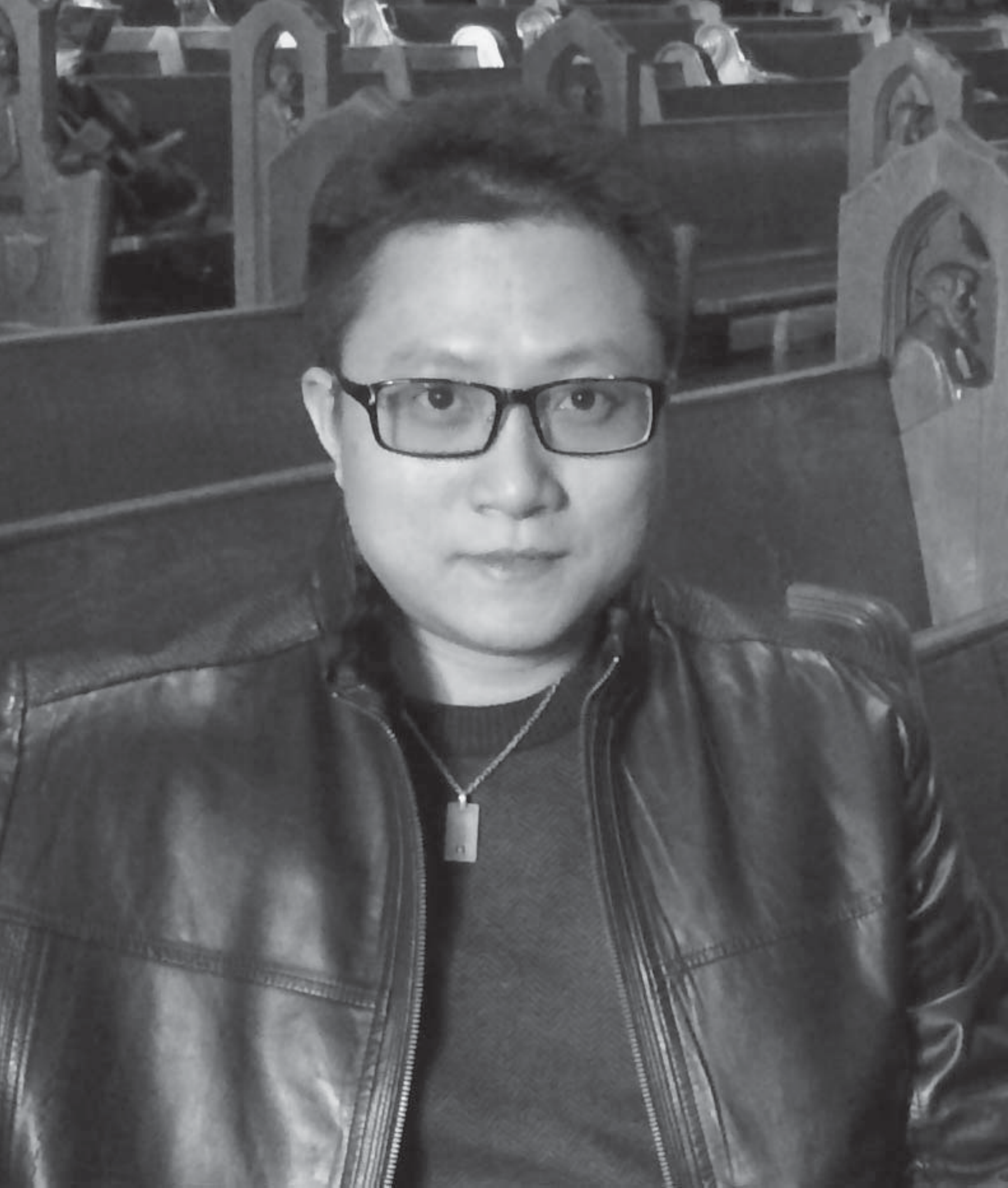}}]{Liang Lin}
	Liang Lin is a full professor of Computer Science in Sun Yat-sen University and CEO of DarkerMatter AI. He worked as the Executive Director of the SenseTime Group from 2016 to 2018, leading the R\&D teams in developing cutting-edge, deliverable solutions in computer vision, data analysis and mining, and intelligent robotic systems.  He has authored or co-authored more than 200 papers in leading academic journals and conferences. He is an associate editor of IEEE Trans.  Human-Machine Systems and IET Computer Vision, and he served as the area/session chair for numerous conferences such as CVPR, ICME, ICCV. He was the recipient of Annual Best Paper Award by Pattern Recognition (Elsevier) in 2018, Dimond Award for best paper in IEEE ICME in 2017, ACM NPAR Best Paper Runners-Up Award in 2010, Google Faculty Award in 2012, award for the best student paper in IEEE ICME in 2014, and Hong Kong Scholars Award in 2014. He is a Fellow of IET.
\end{IEEEbiography}
\end{document}